\documentclass[journal]{IEEEtran}

\usepackage{times}
\usepackage{epsfig}
\usepackage{graphicx}
\usepackage{amsmath}
\usepackage{amssymb}

\usepackage{caption}
\usepackage{subcaption}
\usepackage{array,multirow}
\usepackage{threeparttable}
\usepackage{booktabs}

\usepackage{xcolor}     
\usepackage{soul}       
\usepackage{hyperref}   
\usepackage{comment}    
\usepackage{float}
\usepackage{todonotes}
\usepackage[symbol]{footmisc}

\sethlcolor{white}

\hypersetup{colorlinks=true,linkcolor=blue,urlcolor=blue,citecolor=blue}

\newcommand*\rot{\rotatebox{90}}

\begin{document}

    \def \OurTitle {State Of The Art In Open-Set \\Iris Presentation Attack Detection}
\title{\OurTitle}

\author{Aidan~Boyd*\thanks{* denotes equal contribution.},
        Jeremy~Speth*,
        Lucas~Parzianello*,
        Kevin~Bowyer,
        and~Adam~Czajka
\thanks{All authors are with the Department of Computer Science and Engineering, University of Notre Dame, 384 Fitzpatrick Hall of Engineering, Notre Dame, IN 46556, USA.}
}

\maketitle

\begin{abstract}
Research in presentation attack detection (PAD) for iris recognition has largely moved beyond evaluation in ``closed-set’’ scenarios, to emphasize ability to generalize to presentation attack types not present in the training data. This paper offers several contributions to understand and extend the state-of-the-art in open-set iris PAD. First, it describes the most authoritative evaluation to date of iris PAD. We have curated the largest publicly-available image dataset for this problem, drawing from 26 benchmarks previously released by various groups, and adding 150,000 images being released with the journal version of this paper, to create a set of 450,000 images representing authentic iris and seven types of presentation attack instrument (PAI). We formulate a leave-one-PAI-out evaluation protocol, and show that even the best algorithms in the closed-set evaluations exhibit catastrophic failures on multiple attack types 
in the open-set scenario.
This includes algorithms performing well in the most recent LivDet-Iris 2020 competition, which may come from the fact that the LivDet-Iris protocol emphasizes sequestered images rather than unseen attack types. Second, we evaluate the accuracy of five open-source iris presentation attack algorithms available today, one of which is newly-proposed in this paper, and build an ensemble method that beats the winner of the LivDet-Iris 2020 by a substantial margin. This paper demonstrates that closed-set iris PAD, when all PAIs are known during training, is a solved problem, with multiple algorithms showing very high accuracy, while open-set iris PAD, when evaluated correctly, is far from being solved. The newly-created dataset, new open-source algorithms, and evaluation protocol, all made publicly available with the journal version of this paper, provide the experimental artifacts that researchers can use to measure progress on this important problem.
\end{abstract}


\IEEEpeerreviewmaketitle

    \section{Introduction}
\label{sec:introduction}

\begin{figure*}[ht]
    \centering
    \includegraphics[width=\linewidth]{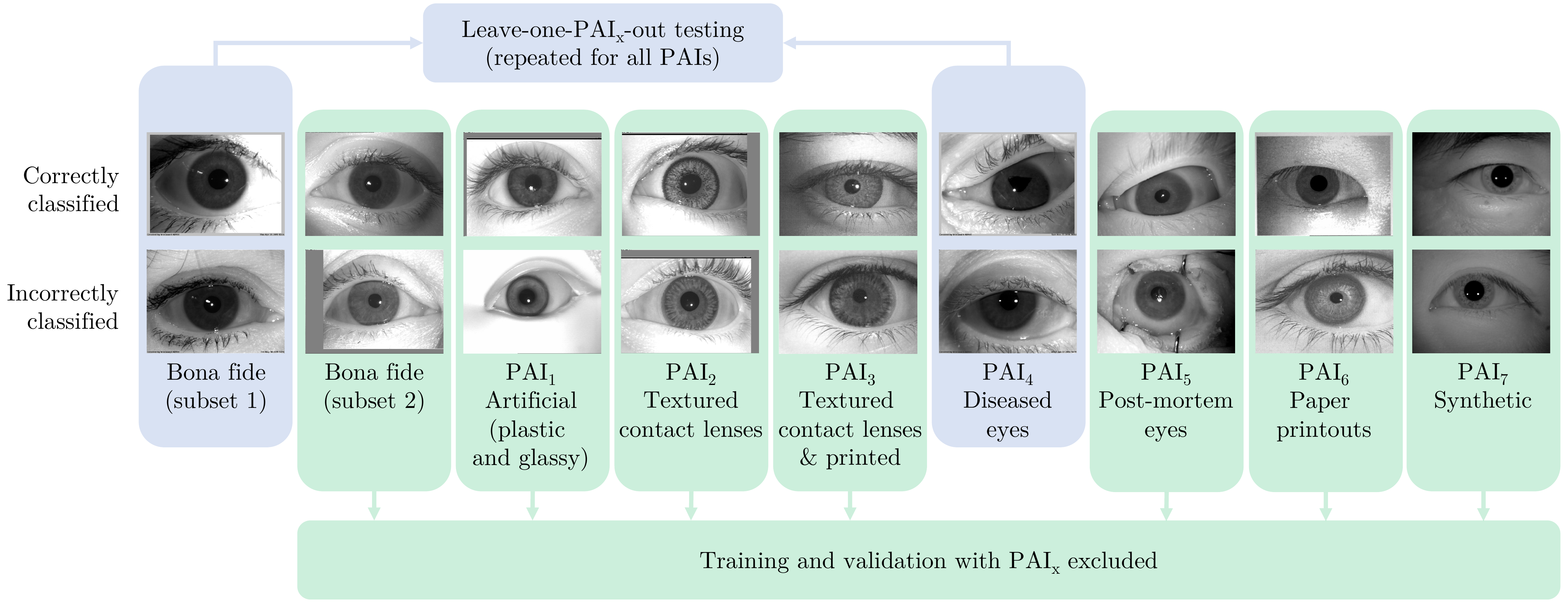}
    \caption{Example images of live irises and seven PAIs (Presentation Attack Instruments) collected from all publicly-available iris PAD datasets and used in this work. The applied evaluation protocol assumes to train the methods on all PAIs except one (``leave-one-PAI-out'' approach) with a subset of live iris samples, and test with that unknown PAI and another (subject- and dataset-disjoint) subset of live samples. This evaluation is repeated $N$ times, where $N$ is the number of all distinct attack types. For illustration, example images classified correctly and incorrectly by the proposed ensemble method, are shown in two rows.}
    \label{fig:attack_examples}
\end{figure*}

\IEEEPARstart{P}{resentation} attacks (PA) are those presentations to a biometric capture subsystem, which aim at driving the system into an incorrect decision. The goal may be either identity {\it concealment} (known in a wider machine learning sense as ``untargeted attacks''), where a subject doesn't want to be recognized, or {\it impersonation} (in a wider context known as ``targeted attacks''), in which the attacker has a knowledge and capabilities to prepare and present artifacts carrying identity information of a specific person. Presentation attack detection (PAD) competing against sophistication of the attacks is a typical arms race, and iris recognition is not an exception here. The first successful attacks on commercial iris recognition systems, using iris images printed on paper, were demonstrated in 2002 \cite{Thalheim_CT_2002}, just one year after successful deployments of iris recognition at several major American, European and Asian airports, and surprisingly can still be observed despite almost two decades of the problem's demonstration. 

Early iris PAD ideas were based on analysing image frequency spectra to locate  frequencies corresponding to printing artifacts \cite{Daugman_Book_1996} and became a successful approach for simple concealment attack attempts using printed contacts.

The arsenal of known iris presentation attacks is now much larger, and includes 
artificial eyes, images displayed on digital (``e-ink'') readers, synthetically-generated images (\eg by modern deep learning tools such as Generative Adversarial Networks), or even cadaver eyes \cite{Trokielewicz_BTAS_2018,Czajka_ACM_2018,boyd2020iris}. 
Modern algorithms, typically deep learning-based, that are trained on data sampled from the same PAI as in the test data demonstrate close-to-perfect PAD accuracy in many recent studies.
This suggests that closed-set iris PAD can be considered as a solved problem.
This is also observed in works testing iris PAD approaches with the LivDet-Iris 2013 and 2015 closed-set benchmarks \cite{Yambay_IJCB_2014,Yambay_ISBA_2017}. The situation changes drastically if not all attack types are known during training. LivDet-Iris 2017 and 2020 were organized partially in an open-set recognition regime, in which not all PAIs included into the test benchmark were disclosed to the participants. As a consequence, the algorithm winning the most recent competition was not able to detect nearly 60\% of attack attempts, although it achieved a favourable bona fide classification error rate of less than 0.5\%.

So what is the true current state of open-set iris PAD? If one would like to deploy an iris PAD algorithm, does the biometric community currently have an effective solution to offer? This paper answers these questions through four important contributions:

\begin{enumerate}
    \item Curation of all 26 publicly-available (at the time of writing this paper) iris PAD datasets, plus an additional 150,000 images released with the journal version of this paper, to create a 450,000-image iris PAD testing benchmark.
    \item Evaluation of all iris PAD solutions proposed to date, with source codes available, on the largest-possible iris PAD benchmark created in (1).
    \item An ensemble classifier built with three selected iris PAD solutions, including one proposed in this paper, demonstrating better accuracy on the most recent LivDet-Iris 2020 benchmark than the competition winner.
    \item A rigorously-defined leave-one-PAI-out testing protocol to evaluate iris PAD in a realistic open-set scenario, as illustrated in Fig. \ref{fig:attack_examples}, with data splits and source codes offered 
    for full reproducibility and future comparisons.
\end{enumerate}

The ensemble classifier proposed in this paper includes a Variational Autoencoder (VAE)-based algorithm, teamed  with two existing solutions based on data-driven (DenseNet) and hand-crafted (Binary Statistical Image Features) approaches. 
This paper presents the most comprehensive evaluation of all existing and open-sourced iris PAD solutions, on the largest-possible iris PAD data corpus, offering an authoritative evaluation of the iris PAD state of the art. 

    \section{Related Work}
\label{sec:related}

In this section, we highlight selected points from recent research in iris PAD, and point the reader to recent surveys for a more comprehensive literature review 
\cite{Czajka_ACM_2018,boyd2020iris}. Iris PAD methods can be classified as {\it static} if they operate on a single sample, or {\it dynamic} if they work on a series of images to extract features related to changes over time. Methods can also be classified as {\it passive}, if they use just an image acquired without special stimulation of the eye, or {\it active}, if the sensor interacts with the subject, such as using lights to prompt pupil constriction \cite{Czajka_ACM_2018}. The PAD methods considered in this work are static and passive. Speed of image acquisition and non-intrusive user interface makes this the most practical category of iris PAD methods.
Also, the accuracy of iris PAD methods using data-driven feature extractors has surpassed that of methods using algorithm knowledge encoded by experts \cite{Sharma_IJCB_2020}. This is a common trend in contemporary machine learning. 

In comparing different methods, we use terminology recommended by ISO/IEC 30107-3:2017 for communicating PAD results.
A presentation (sample) to a biometric sensor is categorized as either \textit{bona fide} -- genuine, belonging to a live human being, and presented without deceptive intention, or \textit{abnormal} -- an artificial object, or a non-conformant presentation of natural characteristics. The metrics following this terminology are CCR, APCER, and BPCER. The {\it Correct Classification Rate (CCR)} is the percentage of classified presentations, whether bona fide or abnormal. The {\it Bona Fide Presentation Classification Error Rate (BPCER)} is the proportion of bona fide presentations incorrectly classified as an attack. Conversely, the {\it Attack Presentation Classification Error Rate (APCER)} is the proportion of attack presentations incorrectly classified as bona fide attempts.

Current approaches to assessing iris PAD differ in the features used, attack types, classifiers, datasets, and evaluation protocols. Sequeira \etal~and Yambay \etal~\cite{Sequeira_TSP_2016, Yambay_IJCB_2017} use different attack types for training and testing and showed that the classification errors greatly increase when these models are evaluated on attack types unseen in training. 
Pinto \etal~\cite{Pinto_DLB_2018} show that the generalization of deep learning methods for presentation attack detection is very limited. Two recent editions of the LivDet-Iris competition \cite{Yambay_IJCB_2017,Das2020} contributed to open-set iris PAD evaluation with their benchmarks that used partially sequestered (from the PAI point of view) test portions of the benchmarks. 

We haven't yet seen many attempts to create iris PAD algorithms that would be truly designed with ``opensetness'' in mind. Sequeira \etal~\cite{Sequeira_TSP_2016} argue that modeling the distributions of spoof samples efficiently, in a scenario where the attack type is uncertain, is not possible. Therefore, they pose the single-class modeling of live samples as an alternative to provide a higher level of certainty against unknown attack classes. In their experiments, a single-class approach resulted in a higher classification error rate than a traditional one. \hl{Their results also vary across attack types and features used, thus being \textbf{inconclusive in terms of effectiveness of one-class classifiers} in the context of iris PAD.} However, the authors consider this a more realistic scenario given the reduction of assumptions about the attack types. Still, they recognize the value of using the knowledge of existing presentation attacks when detecting similar spoof techniques. Along these lines, Yadav \etal~\cite{Yadav_2019_CVPR_Workshops} present a technique for generating synthetic bona fide iris images that can assist the training of a PAD system to learn a representation for live irises, which could help it generalize better to unseen attacks. 
As an additional challenge of creating a generalized PAD classifier, in~\cite{Czajka_ACM_2018} the authors highlight the common pitfalls of biases inherent to the datasets being used for training and evaluation, such as the image capture settings, and a higher probability of observing mascara for women than for men. This can be a problem particularly for approaches using deep-learning, as a result of the limited control over the cues correlating with class labels, and discovered by these approaches during training.

We are not aware of any other work that attempts to evaluate all open-sourced iris PAD algorithms, in a leave-one-PAI-out train/test regime, with 7 different PAIs (\ie all the PAIs available/known today), on a dataset of 450,000 samples.
    \section{Largest Available Iris PAD Dataset} 
\label{sec:dataset}

\subsection{Data sources}

The variety of potential presentation attacks presents an ongoing challenge for iris recognition systems. Assessing how well the designed system processes attack samples requires thorough testing in a scenario when not all attack types are known during method design, and on datasets composed of possibly large number of different PAIs to draw authoritative conclusions regarding generalization capabilities.
To build a dataset that includes the largest possible set of known PAIs, we combined all publicly-available (at the time of writing this paper) benchmarks, and added never-published-before samples collected by our group, to end up with a data corpus of more than 458,790 iris samples representing seven PAIs and a live iris class.
\hl{While we are unable to redistribute the datasets collected outside our institution with this paper, we provide all necessary references for requesting external datasets from the legal owners.}

\hl{Attacks primarily occur at two different points within the iris recognition pipeline: (1) \textbf{presentation} attacks at the sensor when the image is captured; or (2) \textbf{injection} attacks which are digitally inserted prior to feature extraction and matching.} Table \ref{app:full_dataset} and below paragraphs present details and sources for all PAIs in the created benchmark.

\textbf{Artificial irises \hl{(presentation)}:}
The {\it artificial} PAI is composed of 80 plastic eyes from BERC-Iris-Fake \cite{Lee_BS_2006} and 197 images of a glass eye in a Notre Dame dataset \cite{Czajka_TIFS_2017}.

\textbf{Textured contact lenses \hl{(presentation)}:}
The {\it textured contact lenses} PAI consists of contact lenses that give a defined color and texture to the appearance of the iris when viewed normally. Since the lens obscures a large portion of the natural iris, the matching task is rendered more difficult or impossible for these samples. Textured contact lens images were obtained from BERC-Iris-Fake \cite{Lee_BS_2006}, IIITD Iris Contact Lens, \cite{Yadav_TIFS_2014}, LivDet-Iris-2015 Clarkson \cite{Yambay_ISBA_2017}, Clarkson and IIITD-WVU data from LivDet-Iris-2017 \cite{Yambay_IJCB_2017}, and Notre Dame Contact Lens 2013 \cite{Yadav_TIFS_2014} datasets.

\textbf{Textured contact lenses \& printed \hl{(presentation)}:}
This PAI type is images of paper printouts of images in which the person was wearing a textured contact lens.
This is a separate category because it combines effects related to printed images and effects related to textured contact lenses.
In the textured lens category, the images are of a live iris.
In the printouts category, the images are of a printed page of an image where the iris is not wearing a textured contact lens.
When textured contact lenses are the left-out attack type, this category is also left out.
when paper printouts are the left-out attack type, this category is also left out.
When this category is the left-out attack type, paper printouts and textured contact lenses are also left out.

\textbf{Diseased \hl{(presentation)}:}
The {\it diseased} iris images were obtained from the Warsaw-Biobase-Diseased-Iris dataset \cite{WARSAW_DBs_URL}. These samples represent several ocular pathologies, each of which may exhibit different visible changes to the eye that prevent accurate iris recognition. For simplicity, Trokelewicz \etal \cite{Trokielewicz_BTAS_2015} divide the visual changes into clear pattern, geometrical
deviations, iris tissue impairments, and obstructions seen in 
the anterior chamber of the eyeball.

\textbf{Post-mortem \hl{(presentation)}:}
Near-infrared (NIR) eye images {\it captured after death} were obtained from the Warsaw-BioBase-Post-Mortem-Iris v3 dataset \cite{Trokielewicz_IVC_2020, WARSAW_DBs_URL}. Medical staff obtained images from 42 cadavers ranging from several hours up to 369 hours after demise.

\textbf{Paper printouts \hl{(presentation)}:}
Irises {\it printed on the paper} and then photographed in near-infrared were obtained from BERC-Iris-Fake \cite{Lee_BS_2006}, ATVS-FIr \cite{Galbally_ICB_2012, Galbally_TIP_2014}, IIITD Combined Spoofing \cite{Kohli_BTAS_2016}, LivDet-Iris-2015 Clarkson and Warsaw partitions \cite{Yambay_ISBA_2017}, as well as the LivDet-Iris-2017 Clarkson, IIITD-WVU, and Warsaw partitions \cite{Yambay_IJCB_2017}.

\textbf{Synthetic \hl{(injection)}:}
10,000 {\it synthetically-generated} iris images were obtained from the CASIA-Iris-Synthetic dataset \cite{Wei_ICPR_2008}.

\subsection{Data composition and curation}
Combining all of the collected datasets into a comprehensive collection is not as simple as putting all images together. There are three required curation steps steps applied to this work:
\begin{enumerate}
    \item Removing duplicate images found in multiple datasets.
    \item Retaining only single-channel near-infrared images.
    \item Retaining only of $480 \times 640$ pixel images (ISO-compliant resolution).
\end{enumerate}

The first action item is needed due to the way that many existing benchmarks are created: newer versions of such benchmarks are often supersets of the previously-released datasets. This is the case with, for instance, LivDet-Iris competition test sets. 
This presents challenges in evaluation, since samples occurring more than once are effectively given a higher weight than samples occurring only once. It could happen also that the same samples would appear in both training and testing in cross-dataset experiments, which certainly would skew the results and would end up with a significant underestimation of error rates. Aggregating all of the above datasets gives 800,206 images, but the duplicate image check removed 341,416 images, to end up with a combined dataset of 458,790 unique images.

The second and the third action items are to maintain compliance with ISO/IEC 19794-6, which recommends gray-scale $480 \times 640$ pixel images to be used with near-infrared illumination between 700 and 900 nanometers in commercial iris recognition systems. 
(We did not independently verify the wavelength of near-IR illumination for all image sources.)

\begin{table*}[]
\caption{Number and origin of data samples in the {\bf Combined Dataset} and {\bf LivDet-Iris 2020} benchmark, broken by attack types considered in this study.}
\begin{center}
\label{app:full_dataset}
\begin{tabular}{c|ccc|c}
\toprule
& \multicolumn{3}{c| }{\bf Combined Dataset} & {\bf LivDet-Iris 2020}\\
\textbf{Image Type} & \textbf{Contributing Dataset} & \textbf{\# of Samples}& \textbf{Total \# of Samples} & \textbf{\# of Samples} \\

\midrule
Bona fide& \begin{tabular}[c]{@{}c@{}}
ATVS-FIr \cite{Galbally_ICB_2012}\\
BERC\_IRIS\_FAKE \cite{Sung_OE_2007}\\
CASIA-Iris-Thousand \cite{casia-database}\\
CASIA-Iris-Twins \cite{casia-database}\\
Disease-Iris v2.1 \cite{Trokielewicz_BTAS_2015}\\
ETPAD v2 {\cite{ETPAD_v2_URL}}\\
IIITD Contact Lens Iris \cite{Kohli_ICB_2013}\\
IIITD Combined Spoofing Database \cite{Kohli_BTAS_2016} \\
LivDet-Iris Clarkson 2015 \cite{Yambay_ISBA_2017} \\
LivDet-Iris Warsaw 2015 \cite{Yambay_ISBA_2017}\\
LivDet-Iris Clarkson 2017 \cite{Yambay_IJCB_2017} \\
LivDet-Iris IIITD-WVU 2017 \cite{Yambay_IJCB_2017}\\
LivDet-Iris Warsaw 2017 \cite{Yambay_IJCB_2017}\\
Notre Dame (previously released data)$^*$ \\ 
Notre Dame (new data released with journal paper)$^{**}$\end{tabular}
& \begin{tabular}[c]{@{}c@{}}800\\ 2,776\\ 19,952\\ 3,181\\ 255\\ 400\\ 13\\ 4,531\\ 813\\ 36\\ 3,949\\ 2,944\\ 5,167\\ 217,712 \\ 136,524 \end{tabular} & 399,053 & 5,331 \\ \hline

Artificial & \begin{tabular}[c]{@{}c@{}}BERC\_IRIS\_FAKE \cite{Sung_OE_2007} \\ Notre Dame (previously released data)$^*$ \\ Notre Dame (new data released with journal paper)$^{**}$
\end{tabular} & \begin{tabular}[c]{@{}c@{}}80\\ 91 \\ 106 \end{tabular} & 277 & 541 \\ \hline

Textured contact lenses  & \begin{tabular}[c]{@{}c@{}}
BERC\_IRIS\_FAKE \cite{Sung_OE_2007}\\
IIITD Contact Lens Iris \cite{Kohli_ICB_2013}\\
LivDet-Iris Clarkson 2015 \cite{Yambay_ISBA_2017} \\
LivDet-Iris Clarkson 2017 \cite{Yambay_IJCB_2017}\\
LivDet-Iris IIITD-WVU 2017 \cite{Yambay_IJCB_2017}\\
Notre Dame (previously released data)$^*$\\
Notre Dame (new data released with this paper)$^{**}$\end{tabular}
& \begin{tabular}[c]{@{}c@{}}140\\ 3,420\\ 1,107\\ 1,881\\ 1,700\\ 2,751 \\ 16,373
\end{tabular}
& 27,372 & 4,336 \\ \hline

Textured contact lenses \& printed & LivDet-Iris IIITD-WVU 2017 \cite{Yambay_IJCB_2017}  & 1,899  & 1,899 & --- \\\hline

Diseased & Disease-Iris v2.1 \cite{Trokielewicz_BTAS_2015} & 1,537 & 1,537 & --- \\ \hline

Post-mortem & Post-Mortem-Iris v3.0 \cite{Trokielewicz_IVC_2020} & 2,259 & 2,259 & 1,094 \\ \hline

Paper printouts & \begin{tabular}[c]{@{}c@{}}
ATVS-FIr \cite{Galbally_ICB_2012}\\
BERC\_IRIS\_FAKE \cite{Sung_OE_2007}\\
IIITD Combined Spoofing Database \cite{Kohli_BTAS_2016}\\
LivDet-Iris Clarkson 2015 \cite{Yambay_ISBA_2017}\\
LivDet-Iris Warsaw 2015 \cite{Yambay_ISBA_2017}\\
LivDet-Iris Clarkson 2017 \cite{Yambay_IJCB_2017}\\
LivDet-Iris IIITD-WVU 2017 \cite{Yambay_IJCB_2017}\\
LivDet-Iris Warsaw 2017 \cite{Yambay_IJCB_2017}\end{tabular} & \begin{tabular}[c]{@{}c@{}}800\\ 1,600\\ 1,371\\ 1,745\\ 20\\ 2,250\\ 1,766\\ 6,841\end{tabular} & 16,393 & 1,049 \\ \hline

Synthetic & CASIA-Iris-Syn V4 \cite{Wei_ICPR_2008} & 10,000 & 10,000 & --- \\ \hline

Displayed on e-ink device & --- & --- & --- & 81 \\

\midrule
{\bf All combined} & & & {\bf 458,790} & {\bf 12,432}\\
\bottomrule

\end{tabular}
{\footnotesize \vskip1mm ~$^{*}$ University of Notre Dame has released so far 238,266 iris images (bona fide and various PAIs).\\ All previously released data sets are listed, and a copy can be requested at \url{https://cvrl.nd.edu/projects/data} \vskip1mm ~$^{**}$We release 153,003 new iris images with the journal version of this paper, previously never published. This contributes to 391,269 of all iris samples (bona fide and PAIs) released by Notre Dame to date. A copy of this new dataset will be possible to be requested at \url{cvrl@nd.edu} upon acceptance of this paper.} 
\end{center}
\label{tab:dataset-refinement}
\end{table*}
    \section{Available Open-Source Algorithms}%
\label{sec:evaluated_approaches}

We evaluate four existing open-source and one newly-designed methods on our combined dataset to determine the current state of iris PAD, both in open-set and closed-set scenarios. Of the selected methods, three use deep learning for end-to-end detection, and two extract features using texture analysis approaches such as local binary patterns (LBP) and Binarized Statistical Image Features (BSIF) followed by traditional classifiers. These open-source methods were identified in~\cite{Fang_TIFS_2020}, as well as a more modern method,  D-NetPAD, in \cite{sharma2020dnetpad}. 
All five methods are further described below.

\subsection{Domain-Aware Convolutional Neural Network}
Gragnaniello \etal~\cite{Gragnaniello_SITIS_2016} proposed the Domain-Aware Convolutional Neural Network to incorporate prior information on the nature of many spoofing attacks into the CNN training process. Several spoofing modalities are most clearly visible in the high-frequency spectrum. Standard binary cross-entropy for training does not necessarily encourage the network to attend to such frequencies, so they added regularization terms to learn high-pass solutions within the network. The network was trained from scratch for 120 epochs in each of the experiments presented in this paper. The final model for testing was selected based on the highest accuracy on the validation set. This method is further referred to as {\it DACNN}.

\subsection{Regional Features-Based Iris PAD}
Hu \etal~\cite{Hu_PRL_2016} used the relationship between features in neighboring regions to define a method for robust iris PAD. Up to that point, most PAD approaches used low-level pixel features, so they were motivated to define higher-level features by defining a relational measure between local regions. After rescaling the input images to $300 \times 400$ pixels and segmenting the iris, they use a spatial pyramid model to extract features from a coarse-to-fine scale and apply their relational measure via convolution. Next, to build the models they use multiple low-level features such as Local Binary Patterns (LBP), Local Phase Quantization (LPQ), and intensity correlogram. Due to the size of our dataset, memory constraints while applying the classifiers forced us to use Principal Component Analysis (PCA) on the concatenated feature vectors to 
reduce the dimensionality to 600 principal directions. We refer to this method throughout the paper as {\it RegionalPAD}.

\subsection{Open Source Iris PAD Baseline}

The Open Source Iris Presentation Attack Detection~\cite{McGrath2018} baseline utilizes multiple BSIF-based (Binary Statistical Image Features) filters as its feature extractor and a set of standard classifiers (\ie~Support Vector Machine, Random Forests and Multilayer Perceptron) to make predictions. The final decision is given by a majority vote of the classifiers' decisions. Since the selection of individual classifiers in \cite{McGrath2018} is dataset-dependent, we repeated the selection of optimal classifier ensemble in the OSIPAD method by ranking them by average classification error rate (ACER) calculated as the mean of APCER and BPCER. This ranking strategy differs from the one in the original paper (based on CCR) and better addresses imbalanced representations of various attack types in the combined dataset.  In our reproduction of this algorithm, the classifiers included into the ensemble had their hyperparameters optimized on 1\% of the data used for training of other methods. We refer to this method as {\it OSIPAD}.

\subsection{D-NetPAD}%
\label{sec:dnet}

Employing the popular DenseNet-121 architecture~\cite{Huang_2017_CVPR}, Sharma and Ross present D-NetPAD~\cite{sharma2020dnetpad}. D-NetPAD shows strong generalization across a variety of PAIs, datasets and sensors. They  use visualization techniques to show that the approach produces interpretable class activation maps. This led to the conclusion that their methodology performs well on challenging scenarios while maintaining explainability and interpretability. Image pre-processing for this method involves segmentating  the iris using a SegNet based approach~\cite{Trokielewicz_IVC_2020} followed by cropping and resizing to $224 \times 224$ pixels. 

The authors propose to evaluate the performance of the D-NetPAD in terms of True Detection Rate (TDR) at a False Detection Rate (FDR) of 0.2\% \cite{sharma2020dnetpad}. On their proprietary training set, the acceptance threshold was found to be 0.4, where 0 corresponds to a bona-fide sample and 1.0 to a presentation attack. Due to the class imbalance in our dataset, we evaluate two thresholds. First is the threshold at which the FDR is 0.2\% on our validation set. This will be denoted as D-NetPAD1. The second is using the threshold 0.4, as used in the original paper. This will be denoted as D-NetPAD2. Evaluating this approach using two thresholds allows us to show how thresholds set on different validation sets can influence generalization capabilities.

\subsection{Variational Autoencoder}

\begin{figure}[ht]
    \centering
    \includegraphics[width=\linewidth]{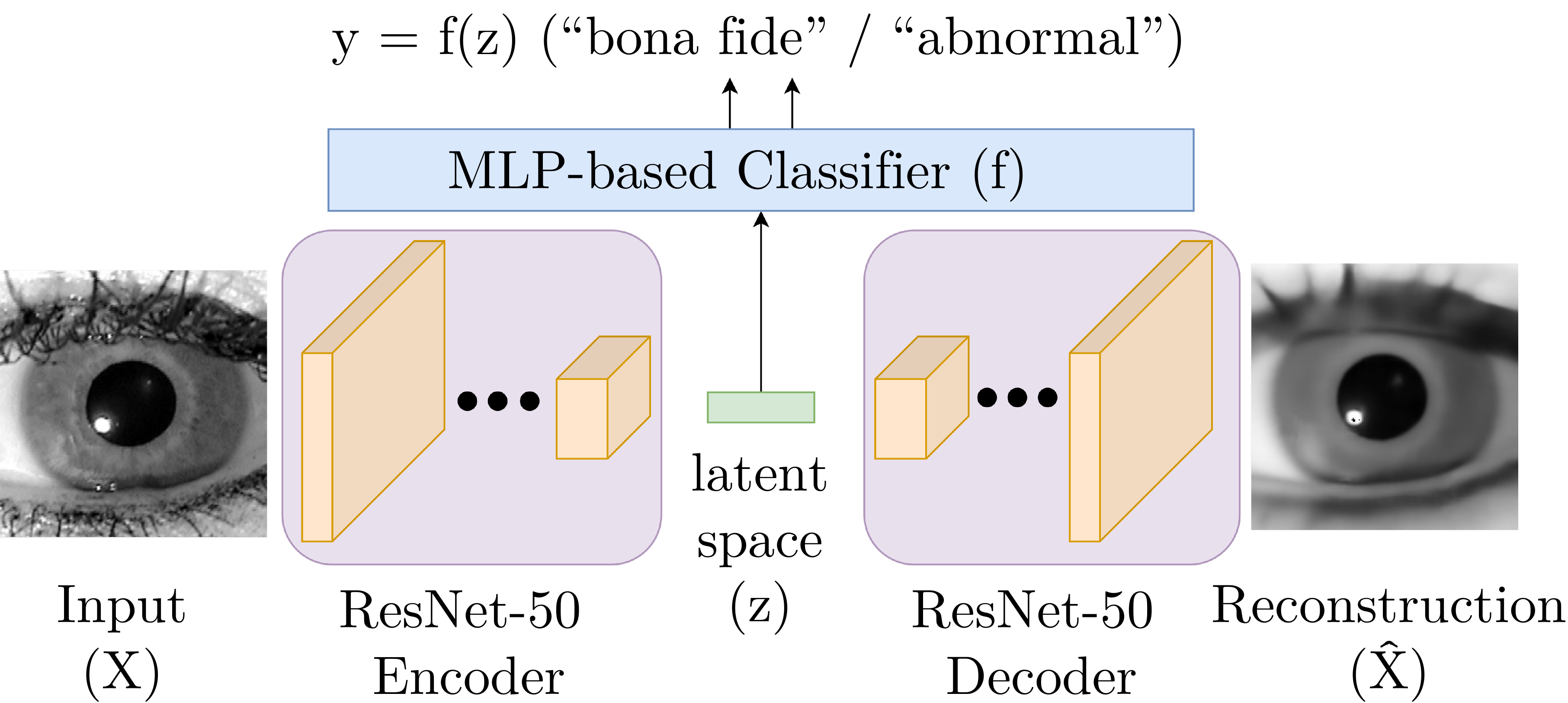}
    \caption{The variational autoencoder is trained to reconstruct bona-fide irises with low error. The latent $\mu$ vectors in $z$ are then used as input features for a multilayer perceptron to predict whether an image is an abnormal sample.}
    \label{fig:vaepad}
\end{figure}

This paper also proposes a new approach to iris PAD based on a variational autoencoder (VAE), which seeks to learn a robust latent representation of live irises.
The VAE consists of a ResNet50 \cite{He_CVPR_2016} encoder and decoder. Our pipeline consists of two steps: (1) we first train the VAE to accurately reconstruct live iris images, and (2) we train a multilayer perceptron (MLP) on the latent $\mu$ vector for binary classification of bona fide or abnormal samples, as shown in Fig.~\ref{fig:vaepad}. 

The first step of training the VAE for reconstructing authentic irises aims to shape the latent space for live samples. The lower-dimensional representation, $z$, must contain relevant information for the decoder to accurately reconstruct the live input images. During training we use the following loss:
\begin{equation*}
    L_{R}(X, \hat{X}) =  -c\,\mathcal{D}_{KL}(Q(z | X)\,||\,P(z)) + \mathrm{MSE}(X, \hat{X}),
\end{equation*}
where $Q(z|x)$ is the approximate posterior, $P(z)$ is the prior distribution of the latent representation $z$, and $\hat{X}$ is the reconstructed image. The left side is the traditional Kullback-Leibler divergence used for regularization, and the right side is for minimizing reconstruction errors. During minibatch training, we use $c=b/N$, where $b$ is the batch size and $N$ is the number of samples in the training set. Note that only live samples are used while training the model for reconstruction.

Next, the weights of the VAE are fixed, and a MLP is created to take the $\mu$ vectors as input to predict whether an input image is an attack. The basic MLP consists of 128, 64, and 2 neurons for the input, hidden, and output layers, respectively. The MLP is trained with both bona fide and abnormal images to minimize the cross-entropy loss after applying softmax to the outputs.

    \section{Experiments}%
\label{sec:experiments}

\begin{figure*}[ht]
    \centering
    \includegraphics[width=1.6\columnwidth]{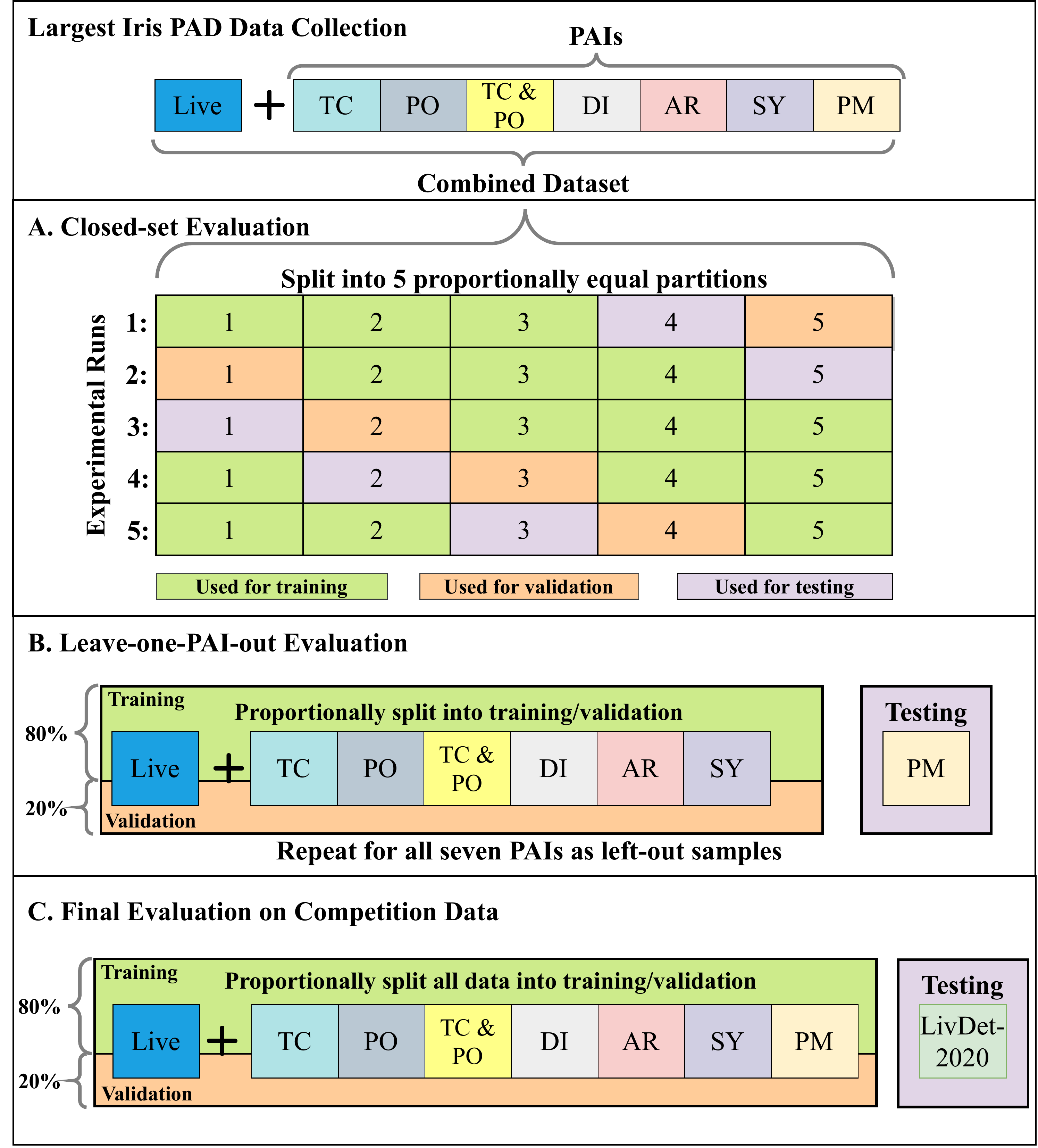}
    \caption{Outline of experiments described in Section IV.  (This Figure is best viewed in color.)  {\it Experiment A is a closed-set evaluation:} all attack types (PAIs) are present in training, validating and testing images. {\bf Experiment B is an open-set evaluation:} one of the seven attack types (PAIs) is held out of training and validation, used only for testing, and the testing cycles through the attack types. {\bf Experiment C follows the LivDet-Iris 2020 competition protocol}, where the combined dataset (458,790 sample) is used for training and validation, and testing is done on the LivDet-Iris benchmark (held-out and independent of the training set). Abbreviations are: TC = Textured Contact Lenses; PO = Paper Printouts; TC\&PO = Paper Printouts with Textured Contacts; DI = Diseased Iris; AR = Artificial Iris; SY = Synthetic Iris; PM = Post-mortem Iris; LivDet-2020 = LivDet-Iris 2020 competition data.}
    \label{fig:experiments}
\end{figure*}

\subsection{Evaluation Protocol Overview}

To assess the performance of the collected open-source iris PAD methods, we complete evaluations under three experimental scenarios. For a closed-set scenario, we split the combined dataset into five equally-sized and equally-distributed portions, with each portion containing all PAIs. 
These splits are then used for closed-set training, validation, and testing. Secondly, each method is evaluated in a leave-one-PAI-out setup, where one of the seven attack types shown in Figure \ref{fig:attack_examples} is removed from training and validation, and held solely for testing. Finally, each method is trained using all available data and evaluated under the LivDet-Iris 2020 protocol~\cite{Das2020}.

Although the data collected for the LivDet-Iris 2020 competition was available to the authors throughout the experimental process, for fair evaluations this benchmark was left out of all experimental training and validation processes in this study, and held out only as the test set. In particular, all references to the collected combined dataset, as described in Section \ref{sec:dataset}, do not include LivDet-Iris 2020 competition data. Figure \ref{fig:experiments} presents graphical representation of all experimental scenarios, and subsections below provide details on each of the evaluation regimes.

\subsection{Closed-set Evaluation}

Closed-set evaluation refers to 
all attack types (PAIs) and sources that the model is tested on also being present in training and validation. 
In general, closed-set performance is high because the variance between the training, validation and testing data is low.

Figure~\ref{fig:experiments}(A) shows how this experiment was executed. 
As well as each split containing approximately the same number of samples, each of the seven PAIs are stratified such that equal numbers of each appear in all splits. Three splits are used for method training, one is used for validation and the last is used for testing. A circular rotation is then performed on the splits resulting in 5 possible train/validation/testing combinations. All methods are evaluated on the same combinations of splits. The final result is the average across all five combinations, along with the error standard deviation. 
Note that the entire training, validation and testing processes were repeated 5 times with all methods. This is a time-consuming and exhaustive process, but allows to calculate fair estimates of how the accuracy's variability depends on the whole design process, not only on the testing sub-process.  

\subsection{Leave-one-PAI-out Evaluation}

In the leave-one-PAI-out scenario, the test set is focused on an unknown attack type that is not present in the training or validation data.  Figure~\ref{fig:experiments}(B) outlines the procedure. First, all images of a specific attack type are 
held out. The remaining data is split 80\%/20\% into training and validation, with the training and validation sets containing equal proportions of each remaining attack type. 

Each method is then designed (including training in case of neural network-based approaches) on the training data, and validated on the validation set. The designed model is then applied to the left-out testing data, which is composed exclusively of one unseen PAI. 
So that the test accuracy is computed for unseen bona fide samples as well as the left-out attack type, 
the bona fide images from the LivDet-Iris 2020 benchmark, collected independently of the data present in the combined dataset, are used in testing. 
This does not break competition protocol as no results in the leave-one-PAI-out evaluation were taken into account for the final evaluation on the complete LivDet-Iris 2020 dataset (detailed in the next subsection).

The goal of this experimental scenario is to explore the generalization capabilities of the studied methods. If a method performs well in a closed-set scenario but poorly on unseen data, it means the learned image representations do not generalize well to new unseen data. 

\subsection{LivDet-Iris 2020 Evaluation}

The third scenario analyzes performance in comparison to the submissions to the most recent LivDet-Iris competition. LivDet-Iris 2020 was an international competition on iris PAD held at IJCB 2020 \cite{Das2020}, open to all interested teams, including commercial entities.
Contestants were asked to submit algorithms trained using any data available to them.
These algorithms were then tested on newly collected data that was not published before the competition, and unknown to the contestants. These unknown samples were from 5 unique PAIs, some of which may not have been present during training.

This same competition data is used as a held out testing set for the methods evaluated in this study, and shown in Figure~\ref{fig:experiments}(C). 
The entire body of data collected as described in Section~\ref{sec:dataset} is split into a training and validation set in an 80\% / 20\% fashion with equal distributions of samples in each attack type. The methods are then trained and validated with these sets. Once the models are fully-trained, they are applied to the LivDet-Iris 2020 data. Evaluation metrics on the individual attack types are reported as well as a comparison to the results obtained in the competition.

Many of the approaches studied in this paper were originally trained on much smaller datasets, including only one or a few of the PAIs represented in the collected dataset. The goal of this experimental setup is to demonstrate the performance capabilities of the studied methodologies on the most recent iris PAD benchmark, when the largest-possible iris PAD dataset is used in model training and validation. 

\subsection{Goals of an Attack}

\hl{
There are two main goals from an attacker's perspective. Our comprehensive dataset contains PAIs for both attack goals. While we do not directly compare attack goals in our experiments, examining the performance of each PAI gives ample evaluation of their individual risk to iris recognition systems.

\textbf{Concealment:}
The first goal is concealment, where the attacker simply tries to obscure identifying features of their own iris in the probe sample.  Textured contact lenses and diseased irises may primarily be used for concealment.

\textbf{Impersonation:}
The second goal is impersonation, where the attacker attempts to gain entry by presenting a sample that contains features from an enrolled identity. The identity could be targeted for a specific individual, or acquired from any identity within the gallery. Artificial, post-mortem, and paper printouts may carry out impersonation attacks. Note that these PAIs could also perform concealment attacks, as it is the least restrictive of the two goals.}
    \section{Results}%
\label{sec:results}

\subsection{Closed-set Evaluation}

\begin{table}[!htb]
    \caption{Closed-set Evaluation Results.}%
    \label{tab:closed-metrics}
    \resizebox{\columnwidth}{!}{%
    \begin{tabular}{@{}cccc@{}}
    \toprule
    \textbf{Method}             & \textbf{CCR ± 1$\sigma$}  & \textbf{APCER ± 1$\sigma$} & \textbf{BPCER ± 1$\sigma$} \\ \midrule
    \textbf{RegionalPAD}        & 94.24\% ± 0.06\%          & 33.41\% ± 0.52\%           & 1.65\% ± 0.05\%            \\
    \textbf{DACNN}              & 98.81\% ± 0.03\%          & 5.36\% ± 0.67\%            & 0.57\% ± 0.08\%            \\
    \textbf{D-NetPAD1}          & 98.53\% ± 1.25\%          & 11.26\% ± 9.63\%           & \textbf{0.00\% ± 0.00\%}   \\
    \textbf{D-NetPAD2}          & \textbf{99.92\% ± 0.01\%} & \textbf{0.18\% ± 0.05\%}   & 0.06\% ± 0.01\%            \\
    \textbf{OSIPAD}       & 99.80\% ± 0.26\%          & 0.85\% ± 1.16\%            & 0.10\% ± 0.12\%            \\
    \textbf{VAEPAD}             & 99.26\% ± 0.03\%          & 3.15\% ± 0.39\%            & 0.38\% ± 0.05\%            \\ \midrule
    \textbf{Ensemble (poly)}    & 99.97\% ± 0.04\%         & 0.14\% ± 0.19\%            & \textbf{0.01 \% ± 0.02} \%          \\
    \textbf{Ensemble (RBF)}     & \textbf{99.97\% ± 0.04\%}          & \textbf{0.09\% ± 0.13\%}            & 0.02\% ± 0.02\%   \\
    \textbf{Ensemble (sigmoid)} & 98.47\% ± 0.28\%          & 5.27\% ± 0.83\%            & 0.97\% ± 0.20\%            \\ \bottomrule
    \end{tabular}}
\end{table}

As the top portion of Table~\ref{tab:closed-metrics} outlines, when evaluating the individual methods in the closed-set scenario, all methods perform well. 
All but RegionalPAD have greater than 98\% test accuracy, and the highest accuracy is observed for D-NetPAD2. This is unsurprising, as D-NetPAD is based on a modern CNN that has been shown to produce strong classification results. Interestingly, D-NetPAD1 (with the acceptance threshold adapted on validation data) has the best BPCER. However, by examining the corresponding APCER results it can be concluded that the threshold for D-NetPAD1 has overfit to the larger bona fide class. This is why experimentation is conducted on two thresholds in this work. Because D-NetPAD2 has the second-best BPCER as well as the best APCER and CCR, it can be concluded to be the strongest approach in the closed-set experiment.

Although closed-set is an unrealistic deployment scenario, it is important to show that the closed-set iris PAD can be considered as a solved problem. The ensemble classification approaches built in this study, and summarized in the bottom part of Table~\ref{tab:closed-metrics}, demonstrate a slight improvement in CCR over individual methods, but we consider this improvement to be not significant, given increased complexity of the overall PAD approach. These results, however, enable a direct comparison with the methods' evaluation in an open-set scenario, as described in the next subsection.

\subsection{Leave-One-PAI-Out Evaluation}

 \begin{table*}
    \begin{center}
    \caption{Leave-One-PAI-Out evaluation results.}
    \label{tab:loo-metrics}
    \resizebox{\textwidth}{!}{%
    \setlength{\tabcolsep}{3pt}{%
    \begin{tabular}{@{}c c ccc c ccc c ccc c ccc c ccc c ccc c ccc@{}}
    \toprule
    \textbf{\begin{tabular}{@{}c@{}}Left Out\\ Attack\end{tabular}}
    && \multicolumn{3}{c}{\textbf{Artificial}}
    && \multicolumn{3}{c}{\textbf{Textured contact lenses}}
    && \multicolumn{3}{c}{\textbf{\begin{tabular}{@{}c@{}}Textured contact lenses\\ \& printed\end{tabular}}}
    && \multicolumn{3}{c}{\textbf{Diseased}}
    && \multicolumn{3}{c}{\textbf{Post-mortem}}
    && \multicolumn{3}{c}{\textbf{Paper printouts}}
    && \multicolumn{3}{c}{\textbf{Synthetic}}\\
    \cmidrule{1-1} \cmidrule{3-5} \cmidrule{7-9} \cmidrule{11-13} \cmidrule{15-17} \cmidrule{19-21} \cmidrule{23-25} \cmidrule{27-29}
        \textbf{Method}
        && \rot{\textbf{CCR}} & \rot{\textbf{APCER}} & \rot{\textbf{BPCER}}
        && \rot{\textbf{CCR}} & \rot{\textbf{APCER}} & \rot{\textbf{BPCER}}
        && \rot{\textbf{CCR}} & \rot{\textbf{APCER}} & \rot{\textbf{BPCER}}
        && \rot{\textbf{CCR}} & \rot{\textbf{APCER}} & \rot{\textbf{BPCER}}
        && \rot{\textbf{CCR}} & \rot{\textbf{APCER}} & \rot{\textbf{BPCER}}
        && \rot{\textbf{CCR}} & \rot{\textbf{APCER}} & \rot{\textbf{BPCER}}
        && \rot{\textbf{CCR}} & \rot{\textbf{APCER}} & \rot{\textbf{BPCER}}\\
        \midrule

        \textbf{RegionalPAD}~\cite{Hu_PRL_2016}
            && 87.23 & 100.0 & 8.24
            && 20.34 & 95.08 & 0.49
            && 26.34 & \textbf{0.05}  & 99.89
            && 71.74 & 99.15 & 7.17
            && 70.00 & 86.01 & 6.27
            && 67.99 & \textbf{10.45} & 98.31
            && 38.05 & 92.41 & 4.80 \\

        \textbf{DACNN}~\cite{Gragnaniello_SITIS_2016}
            && 93.81 & 79.78 & 2.36
            && 16.96 & 99.05 & 0.84
            && 73.68 & 100.0 & 0.08
            && 77.27 & 99.94 & 0.47
            && 76.52 & 77.16 & 0.73
            && 50.26 & 65.71 & 0.64
            && 41.52 & 88.86 & 1.50 \\

        \textbf{D-NetPAD1}~\cite{sharma2020dnetpad}
            && 95.15 & 98.19 & \textbf{0.00}
            && 16.28 & 99.99 & 0.19
            && 73.73 & 100.0 & \textbf{0.00}
            && 77.66 & 99.80 & \textbf{0.00}
            && 70.58 & 98.85 & \textbf{0.00}
            && 26.41 & 97.51 & \textbf{0.04}
            && 34.80 & 99.95 & \textbf{0.02} \\

        \textbf{D-NetPAD2}~\cite{sharma2020dnetpad}
            && \textbf{97.16} & 38.27 & 0.99
            && 16.36 & 99.80 & 0.69
            && \textbf{73.96} & 97.63 & 0.54
            && 78.26 & 95.71 & 0.41
            && \textbf{86.82} & 42.90 & 0.58
            && 43.10 & 75.10 & 0.92
            && 35.56 & 98.29 & 0.94 \\

        \textbf{OSIPAD}~\cite{McGrath2018}
            && 88.69 & 91.70 & 7.13
            && 16.32 & 99.95 & \textbf{0.15}
            && 73.64 & 99.89 & 0.17
            && 69.64 & 94.73 & 11.80
            && 66.43 & 94.82 & 7.62
            && 24.66 & 99.79 & 0.13
            && 33.12 & 99.17 & 6.30 \\

        \textbf{VAEPAD}
            && 94.67 & 59.21 & 2.53
            && 18.78 & 96.70 & 1.74
            && 73.11 & 95.47 & 2.46
            && \textbf{78.87} & 86.21 & 2.36
            && 80.78 & 60.65 & 1.67
            && 38.71 & 80.92 & 0.94
            && 50.34 & 74.79 & 2.51 \\

        \midrule
        \textbf{Ensemble (poly)}
        && 96.35 & 66.43          & 0.39
        && 16.30          & 99.94          & 0.34
        && 73.73 & 98.63         & 0.49
        && 77.84 & 97.92         & 0.32
        && 81.46          & 61.53         & 0.32
        && 34.60          & 86.56         & 0.34
        && 35.54          & 98.59         & 0.45 \\

        \textbf{Ensemble (RBF)}
        && 96.68          & 58.85      & 0.43
        && 16.30          & 99.91      & 0.49
        && 73.62          & 99.32      & 0.39
        && 77.94          & 97.14      & 0.41
        && 83.37          & 54.85      & 0.43
        && 35.60          & 85.20      & 0.43
        && 35.76          & 98.20      & 0.54\\

        \textbf{Ensemble (sigmoid)}
        && 83.17          & \textbf{6.50}  & 17.37
        && \textbf{27.89} & \textbf{83.76} & 12.27
        && 70.66          & 82.83 & 10.28
        && 70.06          & \textbf{71.11} & 18.06
        && 85.46 & \textbf{11.73} & 15.74
        && \textbf{82.16} & 20.34 & 10.17
        && \textbf{53.00} & \textbf{63.17} & 16.66 \\
        \bottomrule
     \end{tabular}
    }}
    \end{center}
 \end{table*}

Table \ref{tab:loo-metrics} presents the results obtained when training and validating each method followed the leave-one-PAI-out protocol. As there are seven attack types represented in the combined dataset, there are seven sets of results presented. The headers in the top row of the table indicate the attack type that was removed from training and validation and held for testing. As previously stated, to evaluate these models on unseen bona fide data, all 5,331 live samples from the LivDet-Iris 2020 data were used in testing.

Similar trends are observed in all seven leave-one-PAI-out experiments in Table \ref{tab:loo-metrics}. 
In each case, the trained method either (1) classifies many of the attack samples in the testing set as live images, as shown by a high APCER, or (2) classifies many of the live samples as an attack, as shown by a high BPCER. 
Contrasting the results in Table~\ref{tab:loo-metrics} with those in Table~\ref{tab:closed-metrics}, it is clear that the methods can efficiently learn relevant features of classes presented during training and validation. However, these learned features turn out to be attack type-specific, which translates into disappointingly poor accuracy of the same methods on unseen PAIs.

Let's now look closer at results in particular attack types. For two commonly-studied PAIs -- the textured contact lenses and paper printouts -- the biometric community has seen numerous algorithms claiming close-to-perfect performance \cite{Czajka_ACM_2018,boyd2020iris}. However, when samples representing these well known PAIs are removed from training, all open-source methods show an inability to detect the nature of these attacks. 
This is an interesting result as it outlines the shortcomings of these approaches in a real-world scenario. In many works, generalization is claimed by training on one variation of an attack, \ie a certain type of textured contact lens and tested on a different dataset of textured contact lenses from different sources. However, when no global domain knowledge is present in the training set, models struggle to differentiate between attack and bona fide samples. 

In all cases, the attained results in the leave-one-PAI-out evaluation detail that the models in all of the seven experiments fail to generalize to the left-out attack. In the case of textured contact lenses, this result is explainable because in many cases the lenses themselves are designed to mimic genuine iris texture. However, in cases where the attack artifacts are pronounced across larger areas, as in the case of paper printouts, models' failure to find them is worrying. 
For postmortem irises, metal retractors may be visible in the sample and should ideally reveal that it is not a genuine sample. However, only D-NetPAD2 correctly classifies more than half of the post-mortem cases (never seen before). In fact, there are only four cases where the APCER is below 50\%. For D-NetPAD2, this occurs in the artificial and postmortem experiments, and for RegionalPAD in the postmortem and paper printout experiments. For RegionalPAD, though, the corresponding BPCER is close to 100\%, revealing how the model has overfit to the attack category. For the other five experiments, D-NetPAD2 also shows overfitting to the bona fide samples, meaning no singular method attained good results on all unseen attacks.

Concluding, the results presented in Table~\ref{tab:loo-metrics} show that open-set iris PAD is an unsolved problem, where much improvement is needed. True evaluations of the open-set capabilities on unseen attack types for the studied methods on large and diverse datasets have not, to our knowledge, been presented prior to this work.

\subsection{LivDet-Iris 2020 Evaluation}

\begin{table}[tbp]
    \centering
    \caption{LivDet-Iris 2020 evaluation results.}%
    \label{tab:livdet-metrics}
    \resizebox{0.9\columnwidth}{!}{%
    \begin{tabular}{ccccc}
    \toprule
    \textbf{Method}       & \textbf{CCR} & \textbf{APCER} & \textbf{BPCER} & \textbf{ACER} \\
    \midrule
        \textbf{RegionalPAD}  & 59.48      & 66.43        & 6.08    & 36.26        \\
        \textbf{DACNN}        & 80.91      & 32.71        & 0.94    & 16.83         \\
        \textbf{D-NetPAD1}    & 67.54      & 56.82        & \textbf{0.02}   & 28.42       \\
        \textbf{D-NetPAD2}    & \textbf{92.71}      & \textbf{12.08}        & 0.90    & \textbf{6.49}      \\
        \textbf{OSIPAD} & 71.75      & 47.13        & 3.10     & 25.12     \\
        \textbf{VAEPAD}       & 79.29      & 33.76        & 3.34      & 18.55    \\
    \midrule
        \textbf{Ensemble (poly)}    & 84.72          & 26.64         & \textbf{0.15}   & 13.40 \\
        \textbf{Ensemble (RBF)}     & 87.45          & 21.76         & 0.28    & 11.02       \\
        \textbf{Ensemble (sigmoid)} & \textbf{93.14} & \textbf{5.82} & 8.25   & \textbf{7.04} \\
    \bottomrule
    \end{tabular}
    }
\end{table}

As a final evaluation of the studied methods, we apply them to the most recent iris PAD competition scenario.
For this evaluation, all data collected in the combined set is used in training and validation. This means all seven attack types are represented in the training data. Training and validation sets are split in an 80\%/20\% fashion with each individual attack type and live samples being proportionally represented in each set.

The LivDet-Iris 2020 competition data represents totally unseen data during training and testing. However, as can be seen in Table~\ref{tab:dataset-refinement}, some of the attack types {\it are} represented in both training and testing, albeit from independent sources. Thus, in this case following the LivDet-Iris 2020 competition protocol can't be considered an open-set evaluation (from the attack types point of view). The intention of this evaluation is to show the effectiveness of the evaluated approaches on data from unknown sources, not necessarily on unseen attack types.

Results for this experiment can be seen in Table~\ref{tab:livdet-metrics}. Here we see much stronger results in comparison to the leave-one-PAI-out experiments. The competition winning solution yielded an \textit{Average Classification Error Rate (ACER)}, which is the average of APCER and BPCER, of 29.78\%. In comparison, when supplied with the largest known collection of iris PAD data, five out of the six studied approaches achieve better results. In the case of D-NetPAD2, the result of 6.49\% bested the competition winner by an astonishing 23.29\%.
These results show the power of large and diverse training and validation sets, especially in the deep-learning based solutions.

Note that all baselines in~\cite{Das2020} were disqualified from the competition by LivDet-Iris 2020 organizers since these institutions had access to test data, which originated from the same source as their train data, unlike the competitors. Hence, it was decided to strictly follow the competition protocol and compare only to the competitors.

    \section{Ensemble Classification}%
\label{sec:ensemble}

\subsection{Ensemble Design}

To address the challenges presented in Section~\ref{sec:results}, an ensemble classifier is built with the goal of achieving greater generalization across  attack types. To compose the ensemble, we selected three classifiers out of those described in Section~\ref{sec:evaluated_approaches} based on their architecture heterogeneity: OSIPAD, an SVM-based classifier that uses BSIF features; D-NetPAD, a CNN-based classifier; and VAEPAD, the novel variational autoencoder-based classifier. The ensemble combines these three classifiers into an SVM-based solution. 

To create this solution, instead of the classifiers returning binary classifications, OSIPAD, D-NetPAD, and VAEPAD were modified to return the assessment of the \textit{likelihood} 
(from 0 to 1) of the input image belonging to the attack class. Then these modified classifiers were combined into an ensemble and SVM-based score-level fusion was chosen to deliver the final assessment of authenticity of a given image. 

Since OSIPAD is an ensemble classifier on its own, its score represents the ratio of individual classifiers that evaluated the image as an attack. For VAEPAD and D-NetPAD, their scores are the values returned by these networks before the application of the decision threshold. 
These three scores are then fed into the SVM, which returns the binary decision of the ensemble. Thus, that scenario implements a classical score-level fusion.

For the ensemble SVM, we explored four different kernels: radial-basis function (RBF), linear, polynomial (with degree 3), and sigmoid. The hyperparameters of these different kernels were not exhaustively explored, as the goal is to capture the generalizability properties of each classifier rather than fine-tune it on a particular dataset. Furthermore, three of these kernels present very similar results across all experiments, as will be shown later.

\subsection{Experimental setup}

Similar to the evaluation of individual classifiers, the ensemble was trained and evaluated in three distinct ways: (i) closed-set evaluation across 5 folds; (ii) leave-one-PAI-out cross-validation; and (iii) on the LivDet-Iris 2020 test set, after training it on the entire combined dataset (sequestered from LivDet competition data).

It is worth observing that there is an extra layer of training on top of existing classifiers.
Now, besides the individual classifier training, an ensemble training stage is necessary. As a result, the original training sets (for each of the three evaluation settings listed above) are used to train the individual classifiers, and their validation sets are used to train different ensemble versions for comparison purposes. The individual validation and ensemble training sets are the same for a given experiment. Similarly, the individual testing and ensemble testing are also the same in a given experiment, thus remaining unseen by all models until the final evaluation.

    \section{Ensemble Classification Results}%
\label{sec:ensemble_results}

\subsection{Ensemble results on the closed set}

As described in Section~\ref{sec:ensemble}, the evaluation of the ensemble model uses four distinct SVM kernels in order to select the best for each task. Table~\ref{tab:closed-metrics} lists the mean and standard deviations of results obtained for each kernel on half of each fold used in cross-validation (with the other half being used to train the OSIPAD ensemble).

The results of the linear kernel were left out of Tables~\ref{tab:closed-metrics},~\ref{tab:loo-metrics}, and~\ref{tab:livdet-metrics} as they present similar behavior to RBF and polynomial kernels, and do not show the best accuracy in any of the tests. RBF and polynomial kernels behave similarly, with low APCER and BPCER, resulting in a near-perfect classification accuracy overall. Additionally, the standard deviation of these metrics across the different folds remains within 1\%, suggesting high stability of these models in a closed-set scenario.

When contrasting the ensemble classifiers against the individual ones in the closed-set experiment, there is a slight average improvement for the best-performing ensemble in terms of CCR (99.97\% compared to the best individual model, D-NetPAD2, achieving CCR=99.92\%).

\subsection{Leave-one-PAI-out}

The closed-set evaluation results show a clear advantage for using RBF or polynomial kernel over sigmoid. However, when evaluated on unseen data, these best-performing classifiers fail to demonstrate the resilience of the models against unknown attack types. 
In order to show how the ensemble would perform in an open-set scenario, the leave-one-PAI-out cross validation was deployed and presented to all SVM kernels. 

Following results in Table~\ref{tab:loo-metrics}, across the seven folds (each with one PAI left out), the results highlight the weaknesses of the three kernels which performed well in the closed-set evaluation. While these kernels never achieve an APCER below 54\%, sigmoid outperforms all of them in this metric across the seven unknown PAIs, while maintaining a BPCER above 80\%. For three out of the seven attack types (artificial, postmortem, and printouts) this ensemble kernel is even able to correctly classify the majority of the unseen attacks presented, at the cost of misclassifying up to 17.37\% of the bona fide images as attacks.

Analyzed on its own, the sigmoid ensemble does not show results that satisfyingly solve the problem of open-set classification of iris PAIs. However, when compared against the individual classifiers working alone (and even the two other ensembles), the sigmoid-SVM-based ensemble outperforms all other classifiers in terms of CCR for three out of the seven folds, and achieves the best APCER on five of these folds. Additionally, in some cases we see a significant advantage in both APCER and CCR, such as on the texture contact lenses and synthetic irises folds.

\subsection{Ensemble performance on LivDet-Iris 2020}

Table~\ref{tab:livdet-metrics} shows the ensemble performances on the LivDet-Iris 2020 competition dataset. Here, the sigmoid kernel presents a similar behavior as seen in the previous evaluations, outperforming the other kernels in terms of APCER and CCR, at the cost of a higher BPCER.

On the comparison of the ensemble against individual classifiers on the LivDet-2020 competition dataset, only the solution employing sigmoid kernel in the SVM wins in terms of the CCR when compared against the best individual classifier in this test (D-NetPAD2). Despite only a slight improvement in this metric, this ensemble model shows the best APCER among all classifiers on the LivDet dataset, supporting the advantage seen in the leave-one-PAI-out experiments.

    \section{Conclusions}%
\label{sec:conclusion}

This study, through a comprehensive open-set evaluation, demonstrates a large gap between {\it de facto} solved close-set iris PAD (in which attack types, not necessarily samples, are known during training), and still-unsolved open-set iris PAD (in which abnormalities in iris images observed during testing were unknown during traning). This paper makes several contributions to assess and push the state of the art in open-set iris PAD, described in the following paragraphs. 

\paragraph{Creation of the largest, open-set-recognition-focused iris PAD assessment benchmark} We have assembled the largest known dataset of publicly accessible, ISO-compliant \cite{ISO_19794_6_2011} iris images.
We focus on ISO compliance in order to ensure relevance and practical impact of results obtained using this dataset.
We focus on publicly-accessible repositories so that results can be reproduced and extended by other researchers.
In this paper we focus on iris PAD, but this benchmark can also be used in research into algorithms for iris recognition, iris image quality, impact of demographics-related information on recognition and PAD performance, and other problems. The supplemental set of 153,003 iris images, and all data splits needed to replicate results in this paper, can be requested at \url{https://cvrl.nd.edu/projects/data} after publication of this work.

\paragraph{Large-scale iris PAD evaluation} We compare results for all currently-known open-source iris PAD algorithms.
We focus on open-source algorithms so that (a) again, results can be reproduced and extended by other researchers,
and (b) there is no question that results presented here do in fact correspond to what is obtained with the published algorithm, rather than a possibly inferior re-implementation.

\paragraph{Empirical demonstration of differences between closed-set and open-set iris PAD}
Our results show clearly that closed-set iris PAD can be considered as ``solved''.
Five previous open-source algorithms achieve a CCR above 98\% and three achieve a CCR of 99.26\% or greater
(see results in Table~\ref{tab:closed-metrics}). Our ensemble algorithm introduced in this paper achieves a slight improvement over the highest CCR of any individual previous algorithm.
It is worth noting that the highest accuracy individual algorithms encompass very different algorithmic approaches.
Thus there are multiple algorithmic approaches through which closed-set iris PAD can be considered as a solved problem.

Unfortunately, but not unsurprisingly, the accuracy results in the open-set context are quite different and much worse.
For the open-set context, the iris PAD problem is quite far from being solved.
None of the six individual open-source algorithms and three variations of our ensemble algorithms achieves more than 28\% CCR with contact lenses as the left-out attack type, or greater than 53\% with synthetic iris images as the left-out type
(see Table~\ref{tab:loo-metrics}).
It is significant there are multiple attack types that seem quite difficult to generalize to, based on learning from the other attack types.
This highlights the danger that an algorithm believed to achieve high accuracy and generalization based on learning from all currently known attack types may still fail spectacularly when it encounters a new attack type.
Future arguments that open-set iris PAD is solved should be based on the lowest left-out accuracy still being at an acceptably high level.

\paragraph{Ensemble classification towards open-set iris PAD} Choosing a best algorithm using closed-set evaluation methodology does not result in selecting an algorithm that will be best in the open-set context.
For the closed-set accuracy results (Table~\ref{tab:closed-metrics}),
multiple algorithms have quite high accuracy, but
D-NetPAD2 and Ensemble (RBF) might be considered the highest-accuracy approaches.
However, for the open-set accuracy results (Table~\ref{tab:loo-metrics}),
Ensemble (sigmoid) has only one left-out type where accuracy falls below 50\%, whereas D-Net-PAD2 and Ensemble (RBF) each have three different left-out types where accuracy falls below 50\%.

\paragraph{LivDet-Iris 2020 evaluation for all open-source iris PAD approaches}
A set of results following the LivDet-Iris 2020 competition protocol
show that our newly introduced iris PAD algorithm significantly outperforms the competition winner and that in fact, multiple algorithms from the literature outperform the the LivDet-Iris 2020 competition.
The strong performance of multiple algorithms from the literature is likely due to the algorithms being re-trained with our large collected dataset.
Unfortunately, while the LivDet-Iris 2020 competition protocol uses a sequestered set of test data, it has images of the same attack types in both test and training data.
The same-attack-type images in the test and training data come from different institutions, but this is not as strong or as realistic of an open-set scenario as true leave-one-attack-type-out.
Thus, results from the LivDet-Iris 2020 competition protocol cannot be taken as indicating trends for the true open-set scenario.

The most difficult attack types clearly are contact lenses and synthetic iris images.
No algorithmic approach achieves greater than 28\% accuracy for contact lenses or greater than 53\% for synthetic iris images.
Better algorithms for detecting both of these attack types are needed.
However, the situation is unlikely to improve rapidly in the near future,
as contact lens manufacturers are likely to pursue a broader variety of more realistic effects, and algorithms for generating synthetic iris images are likely to improve through deep learning approaches.

Concluding, in addition to the contributions of assembling the largest publicly available dataset, comparing all known open-source algorithms, demonstrating why open-set iris PAD cannot be considered as a solved problem, and introducing an ensemble algorithm with improved accuracy in the open-set scenario, the leave-one-attack-type-out methodology used here is one that can be followed in future work even when new datasets and new attack types are introduced.

    \appendices

\ifCLASSOPTIONcaptionsoff
  \newpage
\fi

    \bibliographystyle{IEEEtran}
    \bibliography{main}

\end{document}